\pdfoutput=1

\documentclass[10pt,twocolumn,letterpaper]{article}

\usepackage{iccv}
\usepackage{times}
\usepackage{epsfig}
\usepackage{graphicx}
\usepackage{multirow, multicol}
\usepackage{bbm}
\usepackage{amsmath,amsfonts,amssymb,mathrsfs}
\usepackage{wrapfig}
\usepackage{subcaption}
\usepackage{tablefootnote}

\usepackage{xspace}
\usepackage{makecell}
\usepackage[usenames,dvipsnames,svgnames,table]{xcolor}
\usepackage[bottom]{footmisc}

\newcommand\norm[1]{\left\lVert#1\right\rVert}

\usepackage[resetlabels,labeled]{multibib}
\newcites{Refsupp}{References}

\usepackage[pagebackref=true,breaklinks=true,letterpaper=true,colorlinks,bookmarks=false]{hyperref}

\newif\ifsupp
\suppfalse

\newif\ifarxiv
\arxivfalse
\def\arxivcopy{\global\arxivtrue}

\iccvfinalcopy 
\arxivcopy 

\ifarxiv
\ifsupp
\global\suppfalse
\fi
\fi

\def\iccvPaperID{1} 

\ifsupp
\title{Supplementary Material:\\
CullNet: Calibrated and Pose Aware Confidence Scores for Object Pose Estimation\\ }
\author{Kartik Gupta $^{1,2,3}$, Lars Petersson$^{1,3}$ and Richard Hartley$^{1,2}$\\ 
$^1$ Australian National University, Canberra, Australia \\ 
$^2$ Australian Centre for Robotic Vision \\
$^3$ Data61, CSIRO, Canberra, Australia \\
}
\else
\title{CullNet: Calibrated and Pose Aware Confidence Scores for Object Pose Estimation}
\author{Kartik Gupta $^{1,2,3}$, Lars Petersson$^{1,3}$ and Richard Hartley$^{1,2}$\\ 
$^1$ Australian National University, Canberra, Australia \\ 
$^2$ Australian Centre for Robotic Vision \\
$^3$ Data61, CSIRO, Canberra, Australia \\
{\tt\small kartik.gupta@anu.edu.au, lars.petersson@data61.csiro.au, richard.hartley@anu.edu.au}}

\fi

\ificcvfinal\fi

\begin{document}
\ifsupp
\onecolumn
\maketitle
\thispagestyle{empty}
\appendix
In the supplementary material, we show confidence plots to see calibration effect of CullNet on all classes of LINEMOD \cite{hinterstoisser2012model} dataset. We also show the network architecture of our keypoint proposal network i.e. YOLOv3 \cite{yolov3} in Figure \ref{fig:backbone_arch}.

\begin{figure*}[b]
\centering
          \includegraphics[width=0.8\textwidth]{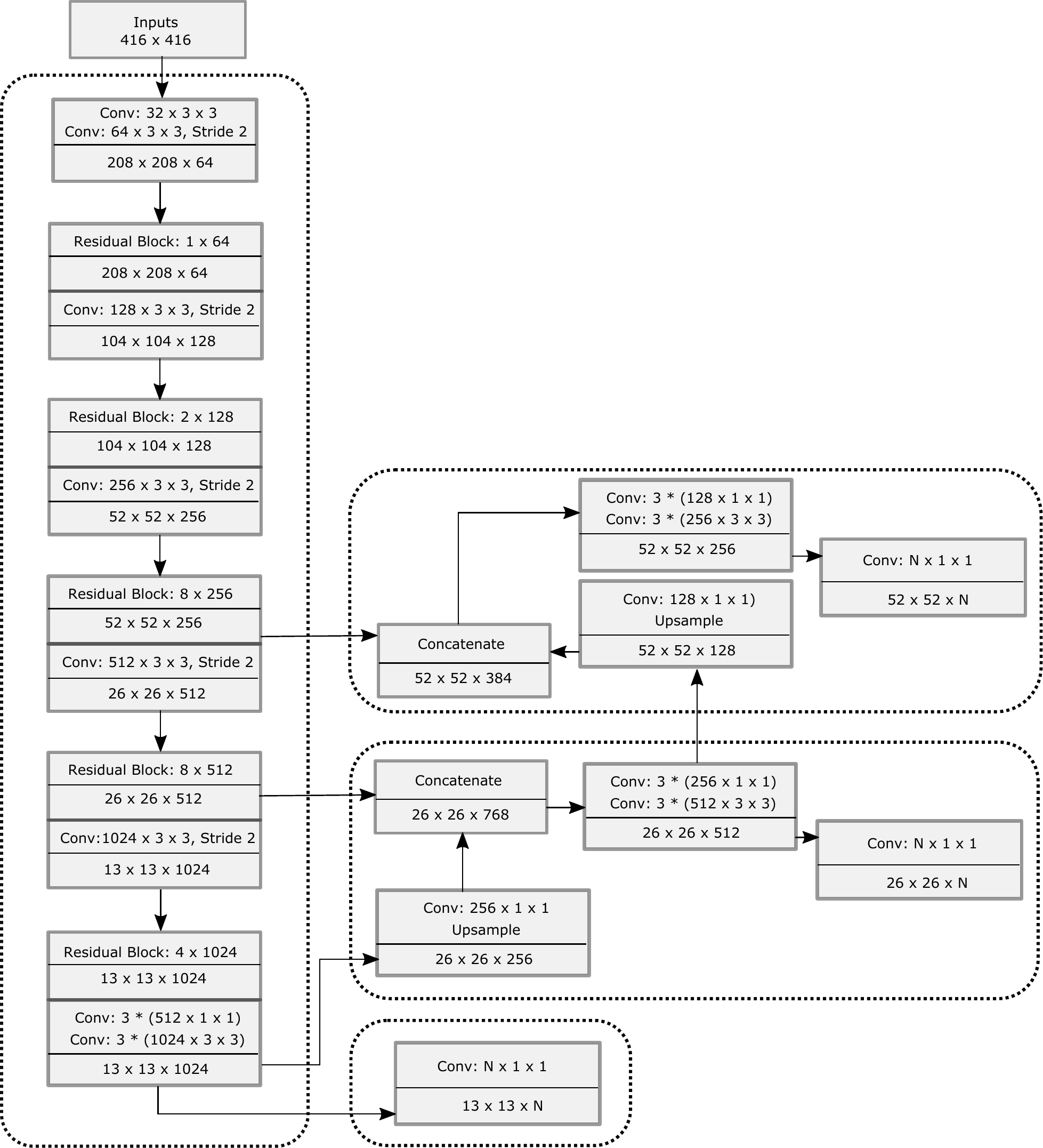}

    \caption{ Network architecture of Keypoint proposal network (YOLOv3 \cite{yolov3}). $N$ denotes the number of outputs from each grid cell.}
\label{fig:backbone_arch}
\end{figure*}

\begin{figure*}[t]
\centering
          \includegraphics[width=\textwidth]{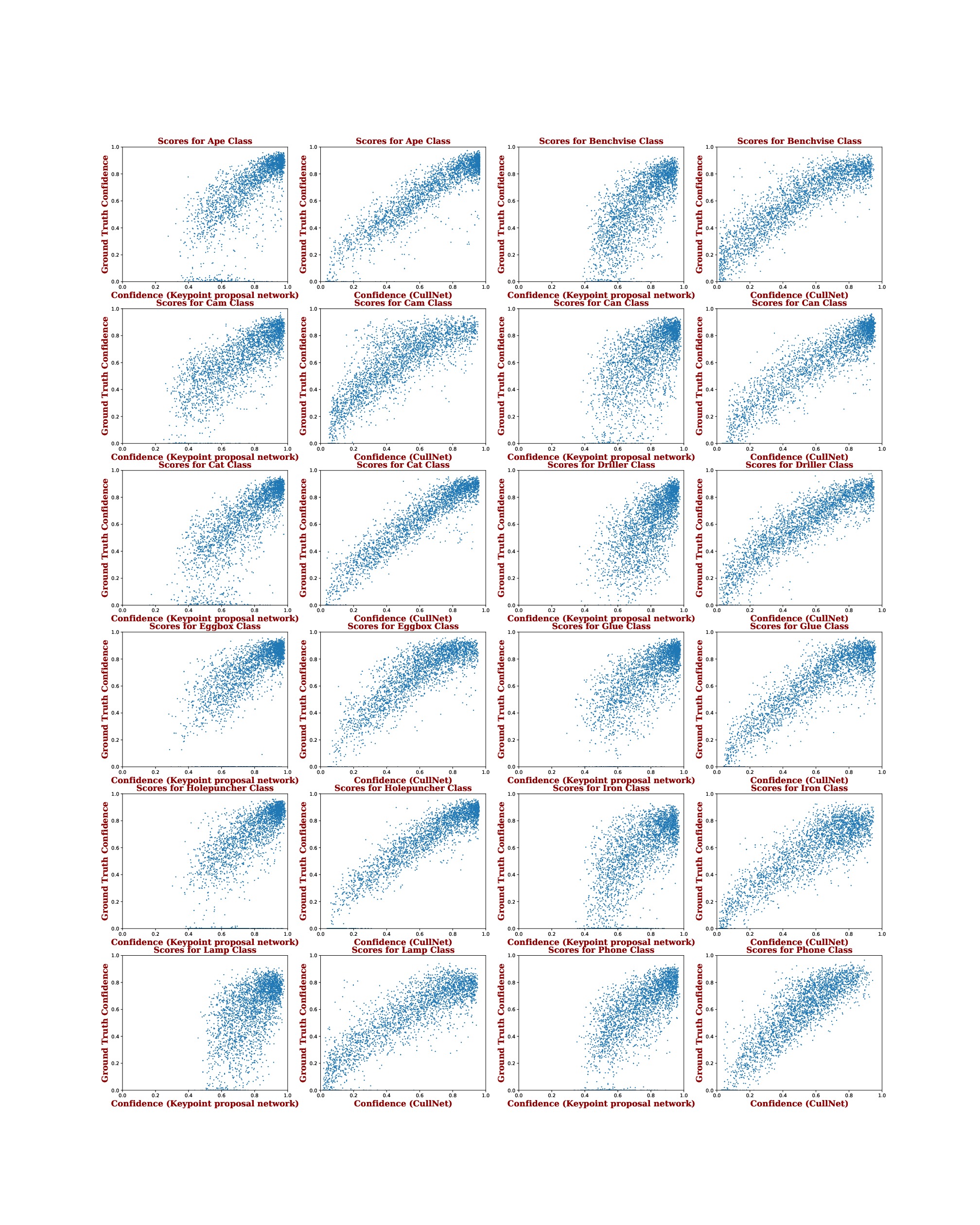}
    \vspace*{-2cm}
          
    \caption{ Comparisons of pose proposal confidence output of Keypoint proposal network and CullNet for different classes of LINEMOD dataset.}
\label{fig:ConfidencePlot_supp}
\end{figure*}

\paragraph{Confidence Plots}
We show comparisons of pose proposal confidence output of Keypoint proposal network and CullNet for all classes\footnote{For 'Duck' class, we show these plots in the main paper.} of LINEMOD dataset in Figure \ref{fig:ConfidencePlot_supp}.

\else

\maketitle
\thispagestyle{empty}

\begin{abstract}
We present a new approach for a single view, image-based object pose estimation. Specifically, the problem of culling false positives among several pose proposal estimates is addressed in this paper. Our proposed approach targets the problem of inaccurate confidence values predicted by CNNs which is used by many current methods to choose a final object pose prediction. We present a network called \textbf{CullNet}, solving this task. CullNet takes pairs of pose masks rendered from a 3D model and cropped regions in the original image as input. This is then used to calibrate the confidence scores of the pose proposals. This new set of confidence scores is found to be significantly more reliable for accurate object pose estimation as shown by our results. Our experimental results on multiple challenging datasets (LINEMOD and Occlusion LINEMOD)  reflects the utility of our proposed method. Our overall pose estimation pipeline outperforms state-of-the-art object pose estimation methods on these standard object pose estimation datasets. Our code is publicly available
 \href{https://github.com/kartikgupta-at-anu/CullNet}{here}.

\end{abstract}
\section{Introduction}
Object pose estimation is crucial for machines to interact with or manipulate objects in a meaningful way. It has applications in various areas such as augmented reality, virtual reality, autonomous driving, and robotics. The challenges to be dealt with are not trivial; background clutter, occlusions, textureless objects, and an often ill-posed formulation where small changes in rotation, translation, or scale can be confused with each other. This paper centers around the particular problem of recovering the six degrees of freedom pose of an object, i.e., rotation and translation in 3D with respect to the camera, dealing with the above-mentioned challenges.

Here, we address the problem of 6-DoF object pose estimation with respect to the camera using an RGB image, and corresponding 3D mesh models of object classes of interest. Specifically, each test image consists of a cluttered environment with a single instance of a textureless object class for which the pose with respect to the camera needs to be estimated. We address this problem on datasets particularly having just a couple of hundred training images with given object pose annotations with respect to the camera. To augment the training data, available 3D mesh models are thus rendered with several different pose variations.

\begin{figure*}[t]
        \begin{subfigure}{0.5\textwidth}
          \includegraphics[width=\textwidth]{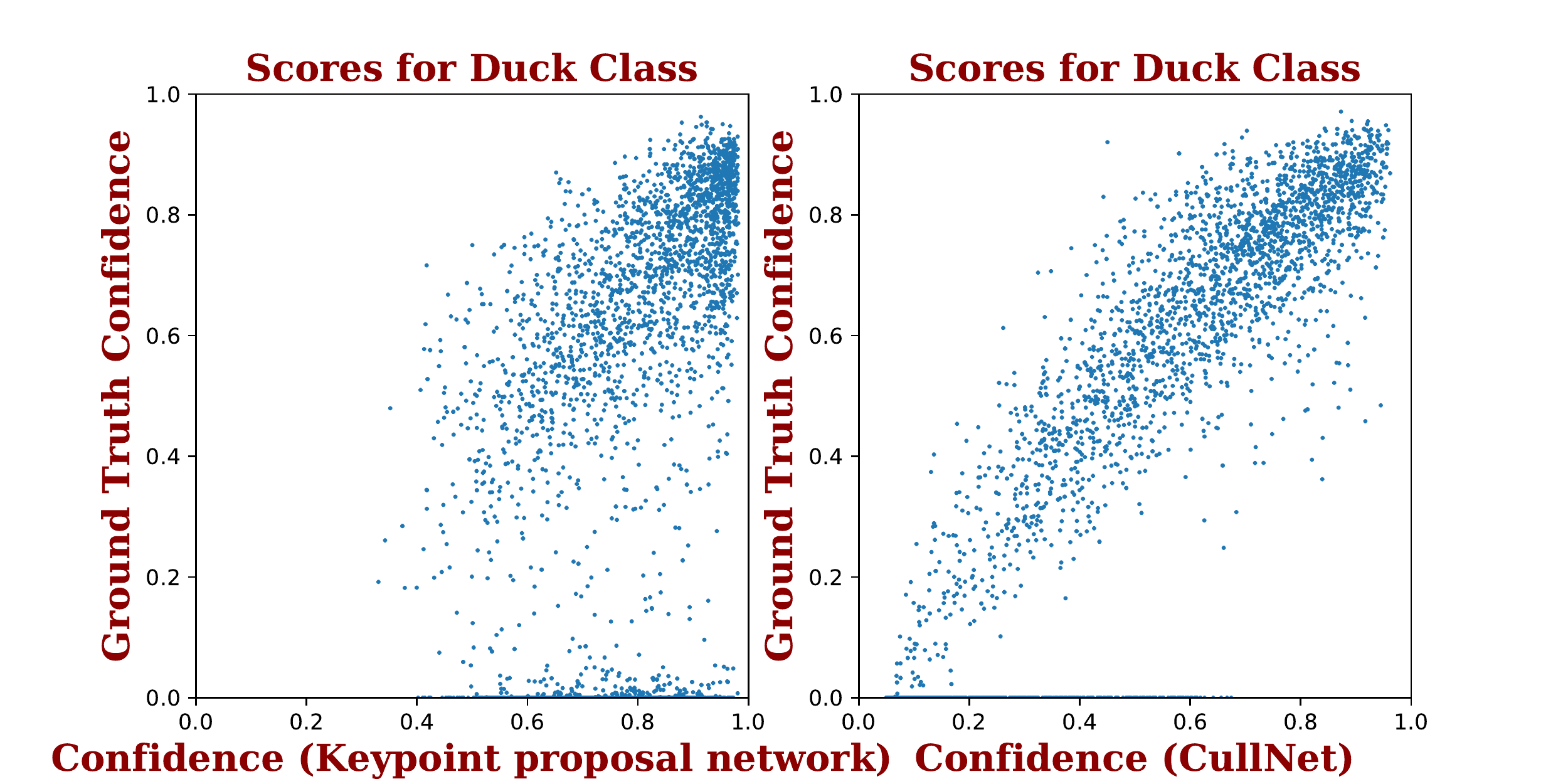}
        \caption{}          
          \label{fig:duck_scatter}
        \end{subfigure}          
        \begin{subfigure}{0.5\textwidth}
          \includegraphics[width=\textwidth]{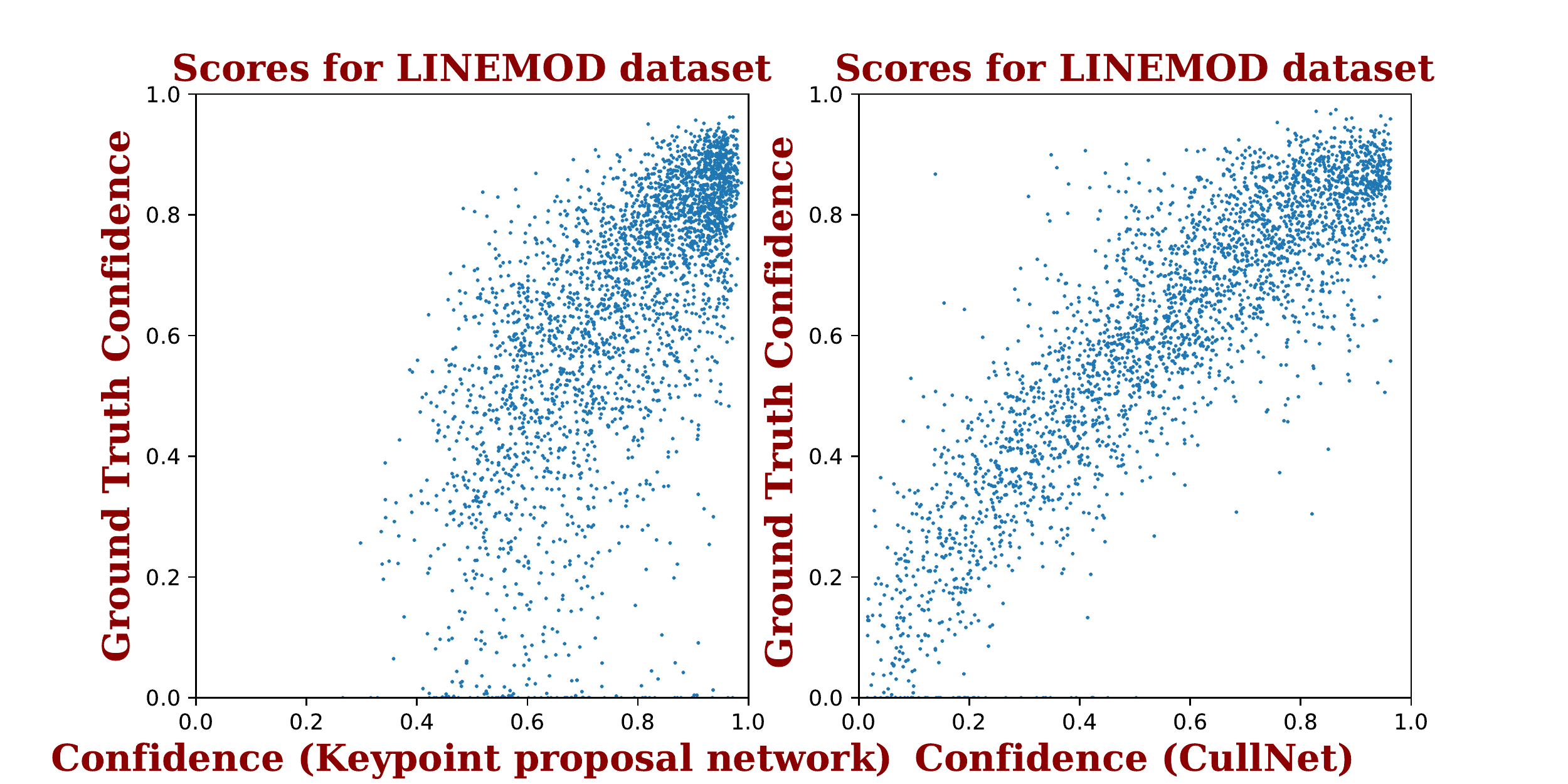}
        \caption{}          
            \label{fig:allclasses_scatter}
        \end{subfigure}          

    \caption{ Comparisons of pose proposal confidence output of the Keypoint proposal network and CullNet. (a) Comparison of confidence scores for the `Duck' class in the LINEMOD dataset. (b) Comparison of confidence scores for all classes in the LINEMOD dataset.}
    \vspace{-6mm}
\label{fig:ConfidencePlot}
\end{figure*}

Overall, this work presents a new approach to predicting several object pose proposals in terms of 2D keypoints, followed by a method to score these proposals. To accomplish this, a fixed number of 3D keypoints are first selected from the object mesh model vertices in the object-centric coordinate system. Given the ground truth pose of the object in each training image, a CNN based on YOLOv3 \cite{yolov3} architecture is trained to predict the 2D projections of these keypoints. Among several sets of keypoints predicted by this CNN, we select the top-$k$ most confident set of keypoints based on their confidence score produced by YOLOv3 and compute the pose for each set of 2D-3D keypoint correspondences using the E-PnP \cite{epnp} algorithm. The object mesh models are then rendered with the predicted $k$ pose estimates to estimate the segmentation masks of the object class of interest. This segmentation mask is tightly cropped along with the input image to form a 4-channel input for our final CNN, i.e. CullNet to find the calibrated confidence scores, used for selecting the most accurate pose proposal. The above two CNNs, in concert, address the object pose estimation problem more accurately because object pose estimation is highly dependent on accurate keypoint proposals. Thus, in this work, many sets of object keypoint proposals are predicted, amongst which an accurate candidate is likely to exist. Via our scoring mechanism, the most accurate keypoint proposal can then be selected as the final prediction.

Recent methods \cite{ssd6d, yolo6d, deepheatmaps} also use deep learning-based methods to predict several pose hypotheses. However, these methods rely on the same backbone network to produce both the pose hypotheses and the confidence measure. Selecting the final pose prediction from this set of hypotheses using the object confidence measures predicted by the same network is undesirable. The reason for this is that the object confidences predicted by the keypoint proposal network do not contain any estimate on how accurate the respective proposed pose is. Thus, we present a new way of re-estimating the object pose confidence measures with an approach that also takes into account knowledge of proposed object poses. We refer to these new scores as calibrated pose proposal confidences. The advantage of using calibrated confidence scores is clearly illustrated in Figure~\ref{fig:ConfidencePlot}. Fig.~\ref{fig:duck_scatter} and Fig.~\ref{fig:allclasses_scatter} compare the distribution of the backbone (keypoint proposal) network or CullNet confidence scores vs. ground truth pose proposal confidence scores for the top-$k$ most confident proposals of all\footnote{To simplify the plot, we randomly sample 3000 confidence outputs from the set of top-$k$ most confident proposals of all test images.} test images of the `Duck' class and all classes together in the LINEMOD dataset respectively. Confidence scores produced by CullNet are more correlated to ground truth confidence scores than scores produced by the keypoint proposal network.

Similar to recent keypoint based methods \cite{yolo6d, bb8, deepheatmaps}, our approach first predicts the 2D keypoints in the RGB images in an end-to-end learning framework. To accomplish this, we employ the backbone architecture of YOLOv3 \cite{yolov3} for the prediction of a set of 2D keypoints. YOLOv3 is one of the fastest object detectors, producing many object proposals for object detection. We amended the original YOLOv3 to predict 2D keypoints rather than object bounding boxes. Then, our proposed method, called CullNet, crops image patches around the top-$k$ most confident keypoint proposals predicted by the backbone network, along with crops of their corresponding, proposed, pose rendered masks. This is used to predict the calibrated confidence measure which can be used either for non-maximum suppression for multi-object pose estimation, or arg-max suppression for single object pose estimation.

Our main contributions are three-fold. \textbf{i}) a new method to calibrate the pose proposal confidences using the knowledge of the corresponding predicted pose, called CullNet, \textbf{ii}) a new keypoint proposal method based on YOLOv3 \cite{yolov3}, which follows a feature pyramid network to predict many sets of keypoint proposals at multiple scales, and \textbf{iii}) an extensive set of evaluations, producing a new state-of-the-art, on the standard benchmark pose estimation datasets: LINEMOD and Occlusion LINEMOD.


\section{Related Work}
Object pose estimation was popularly addressed using keypoint-based methods for a long time \cite{lowe, wagner, rothganger}. However, these methods lack the ability to handle textureless objects as their feature representations require texture information. Recent deep learning based methods try to solve this using CNNs. The solution of the problem requires CNNs to output pose in terms of 3D rotation and 3D translation which has been achieved in different ways.

\paragraph{Direct Pose Prediction}
One way to deal with this is to let the network directly predict the 3D rotation and 3D translation. However, balancing the rotation and translation loss is not trivial as discussed in \cite{posenet}, where they attempt to directly predict rotation and translation vectors for the task of camera re-localization. PoseCNN \cite{posecnn} directly outputs rotation and translation vectors for the object pose estimation by predicting them separately in a multi-stage network. Unlike PoseCNN, which predicts the rotation quaternions, SSD-6D \cite{ssd6d} converts the pose estimation problem into a classification problem by discretizing the views instead of directly predicting the pose. The above-mentioned methods let the network predict the pose from color images directly, which can be difficult for CNNs to achieve, as the CNNs are required to learn all the geometrical knowledge from training data alone.


\paragraph{Keypoint based methods}
Another way to formulate the output of CNNs for object pose estimation is to detect keypoints and then use the Perspective-n-Point (PnP) algorithm \cite{epnp} to estimate the final pose. The works of \cite{deepheatmaps, bb8, yolo6d} achieve significant improvements in pose accuracy on challenging datasets, in particular on textureless objects. 
A key problem in the above methods is inaccurate predicted 2D keypoints. PnP-based pose estimation techniques tend to produce highly perturbed pose estimation results even by small amounts of noise in the predicted 2D keypoints. BB8 \cite{bb8} encounters this problem when predicting a single pose proposal using a CNN on cropped object segments. Due to the noisy regression outputs of CNNs, a single pose proposal often does not result in an accurate one. Also, BB8 is not able to perform the task of object pose estimation in real-time. To this end, Tekin \etal \cite{yolo6d} uses the YOLOv2 object detection network to predict keypoint proposals, but the method lacks an effective way to cull false positives. It uses neighborhood weighted averaging for the keypoints proposals centered around the most confident keypoint proposal. Recently proposed PV-Net \cite{peng2019pvnet} tries to address the problem of partial occlusion in RGB based object pose estimation by regressing for dense pixel-wise unit vectors pointing to the keypoints, which are combined using RANSAC like voting scheme.


\paragraph{Pose refinement methods}
Recent deep learning solutions have also considered techniques for pose refinement from RGB images \cite{deepim, manhardt2018deep} as a way to bridge the gap between RGB and RGBD pose accuracies.
DeepIM \cite{deepim} uses a FlowNet backbone architecture to predict a relative SE(3) transformation to match the colored rendered image of an object using the initial pose to the observed image. Manhardt \etal \cite{manhardt2018deep}  introduce a visual loss that optimizes the predicted refinement of translation and rotation by aligning the contours of the object in a rendered pose with an initial rotation and translation and the scene images. 
The problem specifically targeted in this paper is about culling false positives from several object pose proposals, and such a refinement mechanism can still be used at the end of our pipeline. To the best of our knowledge, there is no work directly addressing the problem of unreliable object pose confidences produced by CNNs. 

Inaccurate object confidences also cause performance degradation in multi-object pose estimation where multiple object pose proposals are predicted for each object. Most state-of-the-art object detection methods~\cite{fasterrcnn, yolov2,ssd}, are dependent on non-maximum suppression (NMS) to cull overlapping, less confident, object proposals. NMS relies on the confidence measure produced by a CNN for a proposal, which is, again, noisy. Our proposed approach addresses the above-mentioned problems associated with the object confidence output of CNNs by calibrating the confidence measures using knowledge of each pose hypothesis. These calibrated confidence predictions can then be used both in single and multi-object pose estimation. 

\section{Approach}
\begin{figure*}
 \centering 
 \scalebox{0.9} 
 {\includegraphics{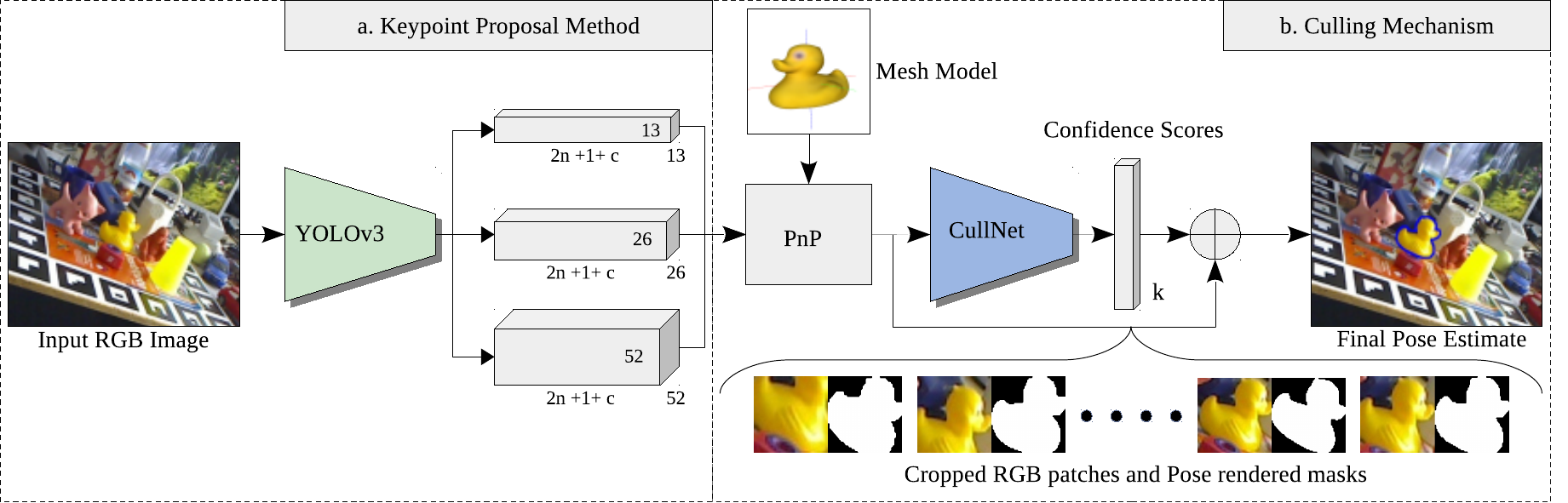}}
 \caption{ Overview of our pose estimation pipeline. Our approach operates in two stages: a) three 3D tensors are outputted using a YOLOv3 based architecture at 3 different scales in the form $2n + 1 + c$ outputs along a spatial grid of each tensor. b) using $k$ sets of 2D keypoint proposals, $k$ pose proposals are estimated using the E-PnP algorithm, then the original image and the pose rendered mask are cropped tightly fitting the rendered mask. Cropped RGB patches concatenated with the corresponding pose rendered masks are passed to CullNet to output calibrated object confidences. Calibrated confidences are finally used to pick the most confident pose estimate.}
 \label{fig:pipeline}
\end{figure*}

In the discussion above, we identify the outstanding issue of overconfident false positives (or inaccurate pose proposals) in current state-of-the-art object pose estimation methods. We address these issues with our proposed object pose estimation pipeline illustrated in Figure~\ref{fig:pipeline}. Our keypoint proposal method is inspired by YOLOv3 \cite{yolov3} which can produce many sets of object keypoint proposals in real-time. Our proposed network, CullNet, produces calibrated object confidences using knowledge of the proposed pose in relation to the observation in the original image. These calibrated confidences can then reliably be used to select a final estimate of the object pose from several object pose proposals.

Using a feature pyramid network \cite{FPN}, the backbone architecture outputs several pose proposals in the form of 2D keypoints. The network is based on the Darknet-53 architecture \cite{yolov3}. One of the crucial advantages of the YOLO network architecture is the gain in speed for object pose prediction, as it is one of the state-of-the-art real-time object detection approaches. The network takes an input image of resolution \(416 \times 416\) and produces outputs at three scales in the form of 3D tensors with spatial sizes \(13 \times 13\), \(26 \times 26\) and \(52 \times 52\) cell grids where each grid point corresponds to a \(2n+ 1 + c\) dimensional vector which includes \(2n\) \(xy\) coordinates of 2D keypoints, \(one\) proposal confidence and \(c\) class scores. In the case of YOLO object detection, the confidence loss is learned based on ground truth IoU overlap between the prediction boxes and the ground truth boxes. Such a formulation of IoU is not easily established in the case of 2D projections of correspondences. We use a confidence function \(c(x)\) proposed in \cite{yolo6d} to assign probabilities to distances between each 2D keypoint in the pose proposals and ground truth 2D keypoints based on some threshold. It is defined as follows:

\begin{equation}
c(x)=
\begin{cases}
\frac{\exp\big(\alpha \big(1 - \frac{D(x)}{d_\textrm{th}} \big)\big) - 1}{\exp(\alpha) - 1},& \text{if} ~~ D(x) < d_\textrm{th}  \\
0              & \text{otherwise}
\end{cases}~.
\label{eq:conffunction}
\end{equation}

The distance $D(x)$ is the Euclidean distance between the predicted 2D keypoints represented by $x$ and respective ground truth 2D keypoint. The confidence function is set to \(0\) for predictions with a distance value greater than or equal to the threshold $d_{th}$. The sharpness of the exponential function is defined by the parameter $\alpha$. In place of the IoU based confidence measure of YOLOv3, the final confidence for each proposal is thus calculated as the mean value of \(c(x)\) over $n$ 2D keypoint predictions.

The backbone network described above predicts \(13 \times 13 + 26 \times 26 + 52 \times 52= 3549\) object pose proposals in terms of \(2n\) \(xy\) coordinates of 2D projected correspondences of object keypoints. In the case of single or multiple instances of an object in the scene, choosing one or many of them is not trivial. 
In object pose prediction, a culling process with inaccurate object confidence scores
often results in culling a better candidate for pose prediction because of being predicted with lower confidence. We thus propose a new confidence calibrating network called CullNet, to predict better confidence measures based on the pose information of each pose proposal. In between the backbone network and CullNet, there is an intermediate processing step to associate pose information with each keypoint proposal as explained below. 

     First, we take the top-$k$ most confident 2D keypoint proposals output by the backbone network and estimate the pose for these $k$ proposals using the PnP algorithm. For each of the $k$ pose proposals, we render binary object segmentation masks. We want to emphasize here that this mask rendering does not require any extensive computation as it does not involve a colored mask. These segmentation masks can simply be calculated by finding the 2D projections of all the mesh vertices of an object. Each rendered mask of proposals is cropped out, tightly around the segmentation boundaries. With the same cropping coordinates, the corresponding RGB patch is formed after cropping the input image. Then, the cropped segmentation mask is concatenated as the fourth channel along with corresponding RGB patch.
     For each top-$k$ most confident proposals, our proposed CullNet takes concatenated RGB patch and mask (\(112 \times 112 \times 4\)) to predict how accurately each pose proposal aligns with the cropped RGB patch. We formulate the ground truth confidence measure for our final output from the proposed CullNet using Eq.~\ref{eq:conffunction}. The Euclidean distance $D(X_{j}, [R_{i}|t_{i}])$ mentioned in Eq.~\ref{eq:conffunction} is
\begin{equation}
D(X_{j}, [R_{i}|t_{i}]) = \norm{K(R_{i}X_j + t_{i}) - K(\hat{R}X_j + \hat{t})}_2.
\label{eq:calibratedconf_gt_1}
\end{equation}

\noindent Here $X_j$ denotes the $j^{th}$ 3D vertex from the object's mesh model, $[\hat{R}|\hat{t}]$ is the ground truth pose, $K$ is the intrinsic camera parameters and $[R_{i}|t_{i}]$ is the $i^{th}$ predicted pose amongst the top-$k$ most confident pose proposals from the keypoint proposal network.  It is important to note here that the final ground truth value for the calibrated confidence i.e. $\hat{C^*_i}$, is the mean over the 2D projections of all $m$ mesh vertices of an object as:
 \vspace{-2mm}
\begin{equation}
\hat{C^*_i} = \frac{1}{m}\sum_{i = 1}^{m}(c(X_i, K[R_{i}|t_{i}])). \label{eq:calibratedconf_gt_2}
\end{equation}

\noindent
CullNet is based on the Resnet50 architecture with the group norm \cite{groupnorm} replacing the batch norm. It takes a 4 channel input of masked out RGB patches. Group Normalization helps in faster convergence of the network with larger batch sizes including patches from the same images having a \(non-i.i.d.\) distribution, that degrades batch norm's statistic estimation \cite{batchrenorm}.

\subsection{Training}
Our complete approach is trained in two stages. First we train the backbone network and then train CullNet using the proposals generated by the backbone network. 
\paragraph{Keypoint proposal network} In the first stage, the backbone network needs to learn prediction of 2D keypoints, confidence scores and class probabilities. The predictions for 2D keypoints are done in the down-scaled size of image coordinates to $13 \times 13$, $26 \times 26$ and $52 \times 52$ respectively. The 2D keypoint predictions are expressed as an offset from the top-left corner of the grid cells. The ground truth confidence scores for the set of 2D keypoints based pose proposal corresponding to each grid cell are calculated using Eq.~\ref{eq:conffunction} where the mean confidence of each set of proposals is calculated as the average over each keypoint confidence. We use a sigmoid function to restrict the predicted confidence score to the range $[0,1]$. We minimize the following loss function to train
our backbone network.
\begin{equation}
\mathcal{L} = \mathcal{L}_\textrm{coord} +  \mathcal{L}_\textrm{conf} +  \mathcal{L}_\textrm{cls}
\label{eq:totalloss}
\end{equation}

\noindent Here, the terms $\mathcal{L}_\textrm{coord}$, $\mathcal{L}_\textrm{conf}$ and $\mathcal{L}_\textrm{cls}$ denote the keypoint, confidence and the classification loss, respectively. We use mean-squared error for the coordinate and confidence losses, and cross entropy for the classification loss. The respective loss functions are formulated as follows for each of the three 3D tensor outputs of the keypoint proposal network:

\begin{equation}
\mathcal{L}_\textrm{coord} = 
\frac{1}{N}\sum_{i = 1}^{S^2}{\mathbf{1}}_{i}^{\text{obj}}\sum_{j = 1}^{n}[(x_{ij} - \hat{x}_{ij})^2 + (y_{ij} - \hat{y}_{ij})^2]
\label{eq:coordloss}
\end{equation}

\begin{multline}
\mathcal{L}_\textrm{conf} = \frac{1}{N}\sum_{i = 1}^{S^2}{\mathbf{1}}_i^{\text{obj}}(C_i - \hat{C}_i)^2 + \\
\frac{1}{M}\sum_{i = 1}^{S^2}(1-{\mathbf{1}}_i^{\text{obj}})(C_i - \hat{C}_i)^2
\label{eq:confloss}
\end{multline}

\begin{equation}
\mathcal{L}_\textrm{cls} = \frac{1}{N}\sum_{i = 1}^{S^2}{\mathbf{1}}_i^{\text{obj}} ( - \mathbf{\hat{y}}_i ^\top \log(\mathbf{y}_i)).
\label{eq:clsloss}
\end{equation}

\noindent where $\mathbf{1}_i^{\text{obj}}$ denotes if the object's centroid keypoint appears in cell $i$, where it is $1$ else it is $0$ and the normalizing constants $N = \sum_{i = 1}^{S^2}\mathbf{1}_i^{\text{obj}}$ and $M = \sum_{i = 1}^{S^2}(1-\mathbf{1}_i^{\text{obj}})$. $C_i$ and $\hat{C}_i$ represent predicted and ground truth confidence scores of the keypoint proposal network. $x_{ij}$, $y_{ij}$ and $\hat{x}_{ij}$, $\hat{y}_{ij}$ denote $xy$ coordinates for $n$ predicted and ground truth keypoints for each set of proposals amongst $S^2$ keypoint proposals, where $S$ varies from $13$, $26$ and $52$ in three different scales. Here, $\mathbf{y}_i$ and $\mathbf{\hat{y}}_i$ represent predicted and ground truth class probability vectors.

\paragraph{Culling Mechanism}
In the final stage of training, CullNet needs to learn a prediction of a calibrated pose-aware confidence measure. We use the sigmoid function to predict outputs of CullNet in the range $[0,1]$. The ground truth calibrated confidences at this stage are calculated based on Eq.~\ref{eq:conffunction}, as an average of the confidence of all 2D projections of mesh vertices at each predicted pose proposal respectively, using Eq.~\ref{eq:calibratedconf_gt_2}. 
For each image, the backbone network passes the top-$k$ most confident object keypoint proposals to the CullNet. Then, pose hypotheses are estimated for each keypoint proposal using the E-PnP algorithm \cite{epnp}. CullNet then uses concatenated cropped RGB image patches and mask renderings as an input (rescaled) for each proposal to produce a confidence measure on how accurate the proposed pose is. We use mean-squared error for the calibrated confidence loss.
\subsection{Inference}
For inference, we first output the top-$k$ most confident keypoints proposals of each object. Then, for each keypoint proposal, the object pose is estimated using the E-PnP algorithm. Based on the predicted pose of the top-$k$ most confident keypoint proposals, tightly cropped object regions in a pose rendered mask and corresponding patches in concatenation are input to CullNet to predict calibrated confidences. Finally, using arg-max on the calibrated confidences outputted by CullNet, we find the estimated pose for the object.

\section{Experiments}
\label{sec:experiments}

\begin{table*}[ht]
   \centering
  \footnotesize
   
    \begin{tabular}{m{.157\textwidth}| p{.02\textwidth} p{.025\textwidth} p{.02\textwidth}  p{.02\textwidth} p{.02\textwidth} p{.03\textwidth} p{.03\textwidth} p{.02\textwidth} p{.02\textwidth} p{.03\textwidth} p{.02\textwidth} p{.03\textwidth} p{.03\textwidth} p{.04\textwidth}} 
&Ape &Bvise &Cam &Can &Cat &Driller &Duck &Box &Glue &Holep &Iron &Lamp &Phone &Avg.\\  
\Xhline{1pt}
	&\multicolumn{14}{c}{\textbf{2D Reprojection-5px}}\\ 
Ours w/ BC & 97.7 & 99.0 & 97.9 &98.9 &98.7 &96.4 &97.0 &98.7 & 98.2 & 99.0 & 97.2 & 95.4 & 95.6 & \textbf{97.7}\\ 
Ours w/o BC & 97.6 & 99.0 & 98.6 & 98.9 &98.6 &96.5 &96.8 &98.7 & 98.3 & 99.0 & 96.0 & 94.7 & 95.1 & \textbf{97.5}\\ 

Tekin \etal \cite{yolo6d} & 92.1 & 95.1 &93.2 &97.4 &97.4 &79.4 &94.7 &90.3 &96.5 &92.9 &82.9 &76.9 &86.1 &90.4\\ 
BB8\cite{bb8} &95.3 &80.0 &80.9 &84.1 &97.0 &74.1 &81.2 &87.9 &89.0 &90.5 &78.9 &74.4 &77.6 &83.9\\
\hline
DeepIM (*) \cite{deepim} &98.4 & 97.0 & 98.9 & 99.7 & 98.7 & 96.1 & 98.5 & 96.2 & 98.9 & 96.3 & 97.2 & 94.2 & 97.7 & \textbf{97.5}\\
BB8\cite{bb8} (*) &96.6 &90.1 &86.0 &91.2 &98.8 &80.9 &92.2 &91.0 &92.3 &95.3 &84.8 &75.8 &85.3 &89.3\\
Brachmann\cite{brachmann2016uncertainty} (*) &85.2 &67.9 &58.7 &70.8 &84.2 &73.9 &73.1 &83.1 &74.2 &78.9 &83.6 &64.0 &60.6 &73.7\\
   	\Xhline{1pt}   	
   	   	&\multicolumn{14}{c}{\textbf{AD\{D$|$I\}-10\%}} \\
Ours w/ BC & 55.1 & 89.0 & 66.2 & 89.2 & 75.3 & 88.6 & 41.8 & 97.1 & 94.6 & 68.9 & 90.9 & 94.2 & 67.6 & \textbf{78.3}\\ 
Ours w/o BC & 34.5 & 79.2 & 71.5 & 85.8 & 71.1 & 89.3 & 39.3 & 86.1 & 87.6 & 70.4 & 85.8 & 73.9 & 63.8 & \textbf{72.2}\\ 
Do \etal \cite{do2018deep} \footnotemark &38.8 &71.2 &52.5 &86.1 &66.2 &82.3 &32.5 &79.4 &63.7 &56.4 &65.1 &89.4 &65.0 &65.2\\ 
Tekin \etal \cite{yolo6d} & 21.6 & 81.8 &36.6 &68.8 &41.8 &63.5 &27.2 &69.6 &80.0 &42.6 &74.9 &71.1 &47.7 &55.9\\ 
BB8\cite{bb8} &27.9 &62.0 &40.1 &48.1 &45.2 &58.6 &32.8 &40.0 &27.0 &42.4 &67.0 &39.9 &35.2 &43.6\\
SSD-6D\cite{ssd6d} & 0 &0.2 &0.4 &1.4 &0.5 &2.6 &0 &8.9 &0 &0.3 &8.9 &8.2 &0.2 &2.42\\
\hline
DeepIM (*) \cite{deepim} & 77.0 & 97.5 & 93.5 & 96.5 & 82.1 & 95.0 & 77.7 & 97.1 & 99.4 & 52.8 & 98.3 & 97.5 & 87.7.0 & \textbf{88.6}\\
Manhardt \cite{manhardt2018deep} (*) &- &- &- &- &- &- &- &- &- &- &- &- &- & 34.1\\
BB8\cite{bb8} (*) &40.4 &91.8 &55.7 &64.1 &62.6 &74.4 &44.3 &57.8 &41.2 &67.2 &84.7 &76.5 &54.0 &62.7\\
SSD-6D\cite{ssd6d} (*) &- &- &- &- &- &- &- &- &- &- &- &- &- &76.3\\
Brachmann\cite{brachmann2016uncertainty} (*) &33.2 &64.8 &38.4 &62.9 &42.7 &61.9 &30.2 &49.9 &31.2 &52.8 &80.0 &67.0 &38.1 &50.2\\
    \end{tabular}
    \caption{ The comparison of accuracies of our method and the baseline methods on the \textbf{LINEMOD} dataset using standard pose evaluation metrics. (*) denotes pose refinement methods. BC refers to bias correction using error modes from train data.}
    \label{tab:linemod} 
\end{table*}


We evaluate our approach on the task of single object pose estimation and show comparisons with the state-of-the-art RGB based object pose estimation approaches.
\subsection{Implementation Details}
\label{subsec:implement_details}
We use Darknet-$53$ pre-trained on the ImageNet classification task as our backbone network. In the Keypoint proposal training, we train only for classification and regression loss for the first $50$ epochs and all losses for the next $50$ epochs. CullNet is trained for $15$ epochs. The sharpness of the confidence function $\alpha$ is set to $2$ and the distance threshold to $30$ pixels. We found $k$ to be best at $6$ keeping the speed-accuracy trade-off in mind. The backbone network has been trained with a batch size of 16 and CullNet with a batch size of 128. We start with a learning rate of $0.001$ for the backbone network using the SGD optimizer and divide the learning rate by a factor of 10 after 50 and 75 epochs respectively. We use a learning rate of $0.01$ for the culling network using the SGD optimizer and divide the learning rate by a factor of 10 after 10 epochs. The number of group norm channels in CullNet are found to be best at $32$. To avoid overfitting, we use extensive data augmentation for training CullNet by randomly changing the hue, saturation, and exposure of the image by up to a factor of $1.5$. We also randomly scale and translate the image by up to a factor of $20\%$ of the image size. During the training of CullNet, we double the number of pose proposals for each image by randomly perturbing the estimated pose from the keypoint proposal network to avoid overfitting. We choose corners and the centroid of the cuboid bounding the object as the 9 keypoints in our experiments (similar to Tekin \etal \cite{yolo6d}).

\subsection{Evaluation Metrics}
\label{subsec:eval_metrics}

We use two standard metrics to evaluate the 6D pose accuracy, namely  2D reprojection error, and the AD\{D$|$I\} metric as used in~\cite{brachmann2016uncertainty,ssd6d, bb8}. 

\textbf{2D Reprojection} measures the  mean distance between the 2D projections of the object's mesh vertices using the ground truth pose and the estimated pose, for each object pose instance. A pose instance is considered correct if the mean distance is less than 5 pixels. 

In contrast, the \textbf{AD\{D$|$I\}} metric measures the mean distance between the
transformed coordinates of mesh vertices using the ground truth pose and the estimated pose for each object pose instance. A pose instance is considered correct if the mean distance is less than $10\%$ of the object mesh model's diameter. To handle rotationally symmetric objects, the mean distance is calculated based on the closest point distance as done in~\cite{bb8}.

\begin{figure*}[!t]
 \hspace*{-0.85cm}
\includegraphics[scale=0.19]{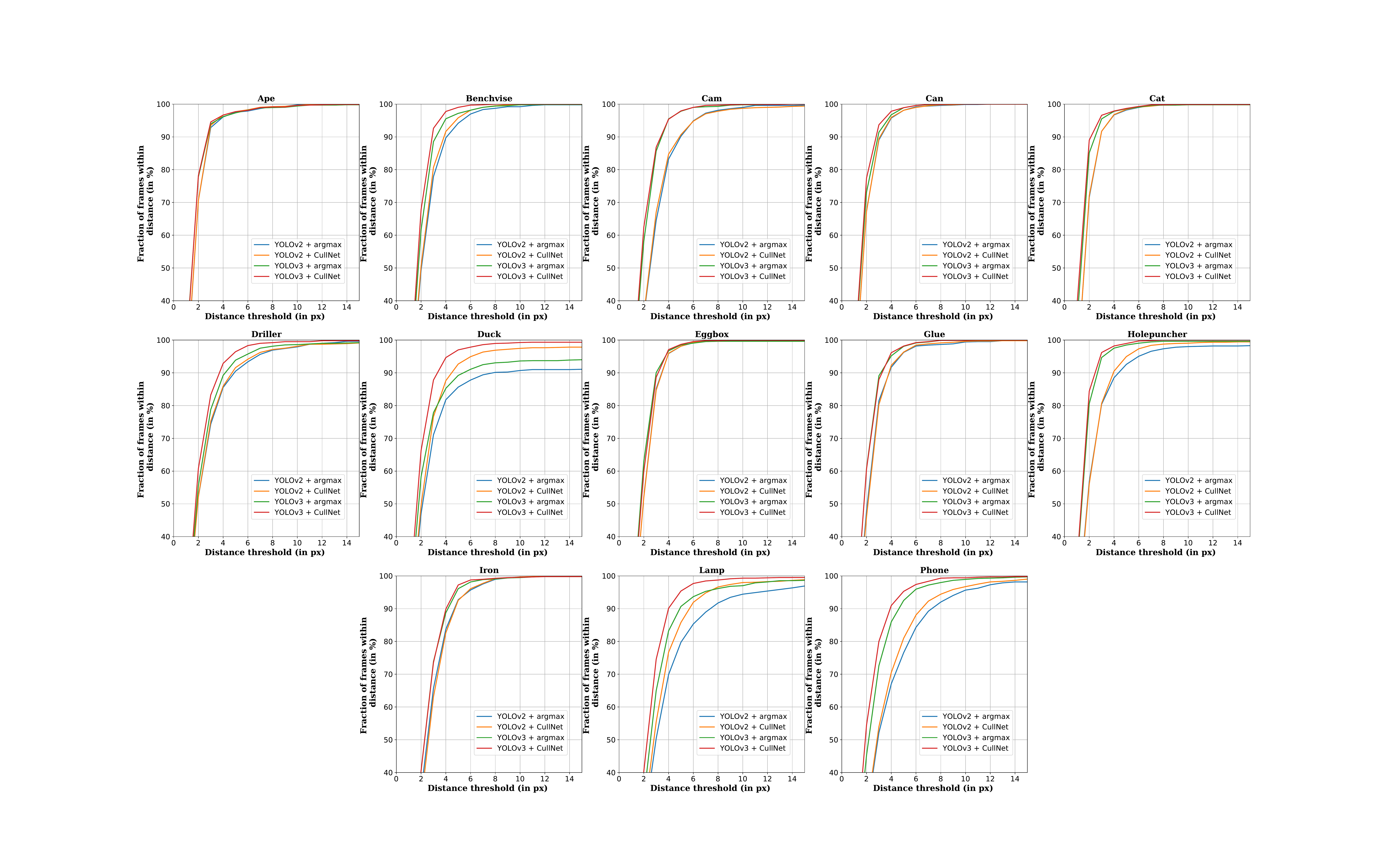}
\vspace{-4mm}
 \caption{Percentage of correctly estimated poses at different 
thresholds of reprojection error (in pixels) for different objects of the LINEMOD
dataset \cite{hinterstoisser2012model}.}
 \label{fig:results_plot_linemod}
\end{figure*}

\subsection{LINEMOD Dataset}

The LINEMOD dataset \cite{hinterstoisser2012model} is a standard benchmark dataset for 6D pose estimation. This dataset is comprised of 13 object classes involving many challenges such as background clutter and textureless objects. Each RGB image has been annotated with only the central object in the scene. We use the same data split for each class as Brachmann \etal \cite{brachmann2016uncertainty} used, with around 200 images for each object in the training set and 1,000 images in the test set. To prevent overfitting, for training we generated synthetic images by rendering objects with uniformly sampled viewpoints with backgrounds randomly selected from the SUN397 dataset \cite{sundata}. To keep the distributions of real and synthetic images the same and also to avoid learning any information from the checkerboard background, we augment the real training images by using the segmentation mask from real images and changing the background from images randomly sampled from the PASCAL-VOC dataset \cite{pascalvoc}.

We show comparisons with competing RGB based object pose estimation methods in Table \ref{tab:linemod}. Our approach outperforms all existing methods comfortably on the 2D-Reprojection metric. It also performs slightly better than the state-of-the-art pose refinement methods on this metric. We want to emphasize the fact that our method, which works using a two-stage pipeline does not use any pose refinement method. Pose refinement methods most often require multiple iterations of refinement along with complete colored renderings of mesh models. Our approach requires only a segmentation mask rendering from the top-$k$ confident pose estimates to calibrate the confidence scores within a single pass through CullNet.
\footnotetext{For this method, results on the 2D Reprojection metric are not available.} 

Our proposed approach also performs better than all existing comparable methods when evaluated on the AD\{D$|$I\} metric. However, the DeepIM \cite{deepim} pose refinement method outperforms our approach on this metric whereas ours perform better on the 2D-Reprojection metric. We investigated this issue which led to the findings that the LINEMOD dataset has many instances of noisy pose annotations due to registration errors between the RGB and the depth image because the pose annotation process was done using ICP on the depth images. A similar observation was also made by Manhardt \etal \cite{manhardt2018deep} evaluating their deep pose refinement method. To partially address this issue, we calculate the error statistics on the LINEMOD training data using the ADD metric from the pose estimated by our final trained network pipeline. We make the histogram plots (using 400 bins) for the ADD error in z-axis after transforming coordinates of mesh vertices using the estimated pose for each object pose instance. Then, we use the modes of training errors along the z-axis for each class as an offset to correct the bias. The offset is added to the translation in the z-axis of all the predicted pose instances by our method, to partially solve the bias problem arising due to noisy annotations. 

\begin{table*}[!t]
\centering
\footnotesize

\begin{tabular}{p{6cm}p{6cm}}
    \centering
    \footnotesize
\begin{tabular}{c|c}
  \multirow{2}{*}{\textbf{Culling Methods}}  & \textbf{2D-Reprojection} \\ 
                  & \textbf{Metric} \\
\Xhline{1pt}

 YOLOv2 + argmax & 91.6 \\ 
 YOLOv3 + argmax & 95.7 \\ 
 YOLOv2 + Top-6 CullNet & 93.4\\
 YOLOv3 + Top-6 CullNet & 97.7\\
\Xhline{1pt}

\end{tabular}&
\raisebox{-0.3\height}{\includegraphics[width=6cm]{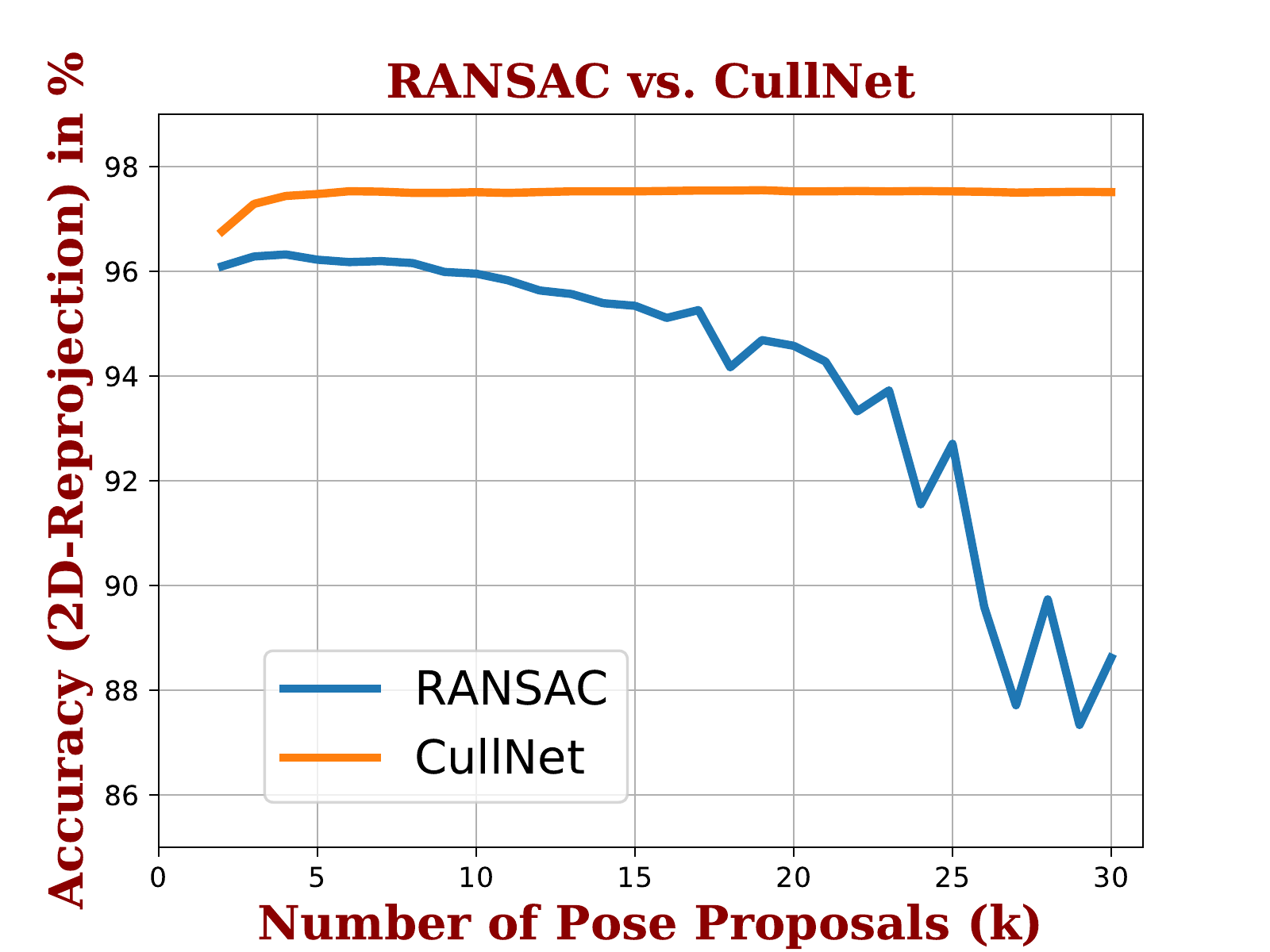}}\\
\vspace{1mm}

\it (a) Accuracy comparisons on LINEMOD dataset using different culling methods on multiple keypoint proposal networks. & 
\vspace{1mm}

\it (b) Robustness of RANSAC vs. CullNet with varying number of top-$k$ most confident pose proposals using YOLOv3 as keypoint proposal network. 
\end{tabular}\\
\vspace{1mm}

\caption{ Ablation studies to show the effectiveness of CullNet on \textbf{LINEMOD} dataset.}
\label{table:ablationstudy}
\end{table*}
\begin{table*}[!t]
    \centering
    \footnotesize
        \begin{tabular}{c|ccccc|ccc}
        & \multicolumn{5}{c|}{\textbf{2D Reprojection-5px}} & \multicolumn{3}{c}{\textbf{AD\{D$|$I\}-10\%}} \\
        
        \multirow{2}{*}{Methods} & BB8 & Tekin & PoseCNN & Jafari & OURS & Tekin & PoseCNN & OURS \\ 
        & \cite{bb8} & \cite{yolo6d} & \cite{posecnn} & \cite{jafari_ipose} & (with BC) & \cite{yolo6d} & \cite{posecnn} & (with BC) \\ 
        \Xhline{1pt}
        ape & 28.5 & 7.01 & 34.6 & 24.2 & \textbf{55.98} & 2.48 & 9.6 & \textbf{21.97} \\ 
        can & 1.20 & 11.20 & 15.1 & 30.2 & \textbf{39.11} & 17.48 & \textbf{45.2} & 24.52 \\ 
        cat & 9.60 & 3.62 & 10.4 & 12.3 & \textbf{34.2} & 0.67 & 0.93 & \textbf{9.77} \\ 
        driller & 0.0 & 1.40 & 7.4 & - & \textbf{29.32} & 7.66 & \textbf{41.4} & 26.11 \\ 
        duck & 6.80 & 5.07 & 31.8 & 12.1 & \textbf{53.46} & 1.14 & 19.6 & \textbf{23.62} \\ 
        eggbox & - & - & \textbf{1.9} & - & 0.17 & - & \textbf{22} & 20.43 \\ 
        glue & 4.70 & 6.53 & 13.8 & 25.9 & \textbf{23.48} & 10.08 & \textbf{38.5} & 28.02 \\ 
        holepuncher & 2.40 & 8.26 & 23.1 & 20.6 & \textbf{72.98} & 5.45 & 22.1 & \textbf{41.4} \\ 
        average & 7.60 & 6.16 & 17.2 & 20.8 & \textbf{38.59} & 6.42 & \textbf{24.9} & 24.48 \\ 
        \Xhline{1pt}
        \end{tabular}
        \vspace{0.2cm}
\caption{ The comparison of accuracies of our method and the baseline methods on the \textbf{Occlusion LINEMOD} dataset. BC refers to bias correction using error modes from train data.}
\label{tab:occlusion}
\end{table*}

\vspace{-2.3mm}
\subsection{Ablation Studies}
\label{sec:ablation studies}
We conduct ablation studies to evaluate the effectiveness of CullNet in comparison to other potential methods for the culling process on the LINEMOD dataset in Table \ref{table:ablationstudy} (a) and Figure \ref{table:ablationstudy} (b). Two such candidate methods are the arg-max selection of the most confident pose proposal and using RANSAC on the top-$k$ most confident pose proposals. 

We evaluate CullNet on top of multiple keypoint proposal networks, namely YOLOv2 and YOLOv3. Our method comfortably outperforms argmax based selection of the most confident pose proposal for both keypoint proposal networks as shown in Table \ref{table:ablationstudy} (a). This clearly reflects the problem of un-calibrated confidence scores in case of argmax based selection in both YOLOv2 and YOLOv3. We also show pose accuracies for all classes of the LINEMOD dataset at varying reprojection error thresholds in the 2D-Reprojection metric in Figure \ref{fig:results_plot_linemod}. These results resonate the effectiveness of CullNet in improving the final pose estimates over a varying range of reprojection error thresholds for the 2D Reprojection metric.

We also show how robust our method is to variations in the number of most confident pose proposals chosen for the culling process in Table \ref{table:ablationstudy} (b). CullNet is shown to be extremely stable to a large number of pose proposals whereas the accuracy starts degrading as $k$ grows in the case of RANSAC. This is related to the fact that our method can differentiate between falsely detected object regions and correct object regions. This property specifically helps in cases where after increasing $k$, we introduce false object proposals such as yellow cup instead of yellow duck.

\subsection{Occlusion LINEMOD Dataset}
Though this work does not attempt to address the problem of partial occlusions in RGB based object pose estimation, it is interesting to see how our approach behaves on such hard examples after training only on the completely un-occluded pose instances. For this, we evaluated our approach on the Occlusion LINEMOD dataset \cite{brachmann2014learning}. This dataset was created by annotating 8 objects in a sequence of 1215 frames from the LINEMOD dataset. This dataset contains challenging cases of severe partial occlusions. We use the same trained models for evaluation on the Occlusion LINEMOD dataset as we use for the LINEMOD dataset.

We show comparisons with state-of-the-art RGB based pose estimation methods on the Occlusion LINEMOD dataset in Table \ref{tab:occlusion}. Our approach outperforms most of the state-of-the-art methods with a huge margin on the 2D-Reprojection metric. It also performs comparably against state-of-the-art on the AD\{D$|$I\} metric. This is an interesting result considering that we do not use any occluded examples during our training process.


\section{Conclusion}
We have introduced a new object pose estimation pipeline based on RGB images only. Our pose estimation pipeline consists of a keypoint proposal network producing several object pose proposals and a new culling mechanism to select the best final pose estimate. We show detailed experimentation on two challenging benchmark datasets where it outperforms state-of-the-art methods. We also show the superiority of our approach to RANSAC and other culling strategies in terms of pose accuracies and robustness against variations in the number of pose proposals.

\fi

\ifarxiv
\clearpage
\onecolumn
\appendix
\begin{center}
     {\bf \Large Supplementary Material} \\ \vspace{.3cm}
     {\bf \Large CullNet: Calibrated and Pose Aware Confidence Scores for Object Pose Estimation}\\\vspace{.2cm}
 \end{center}

In the supplementary material, we show confidence plots to see calibration effect of CullNet on all classes of LINEMOD \cite{hinterstoisser2012model} dataset. We also show the network architecture of our keypoint proposal network i.e. YOLOv3 \cite{yolov3} in Figure \ref{fig:backbone_arch}.

\begin{figure*}[b]
\centering
          \includegraphics[width=0.8\textwidth]{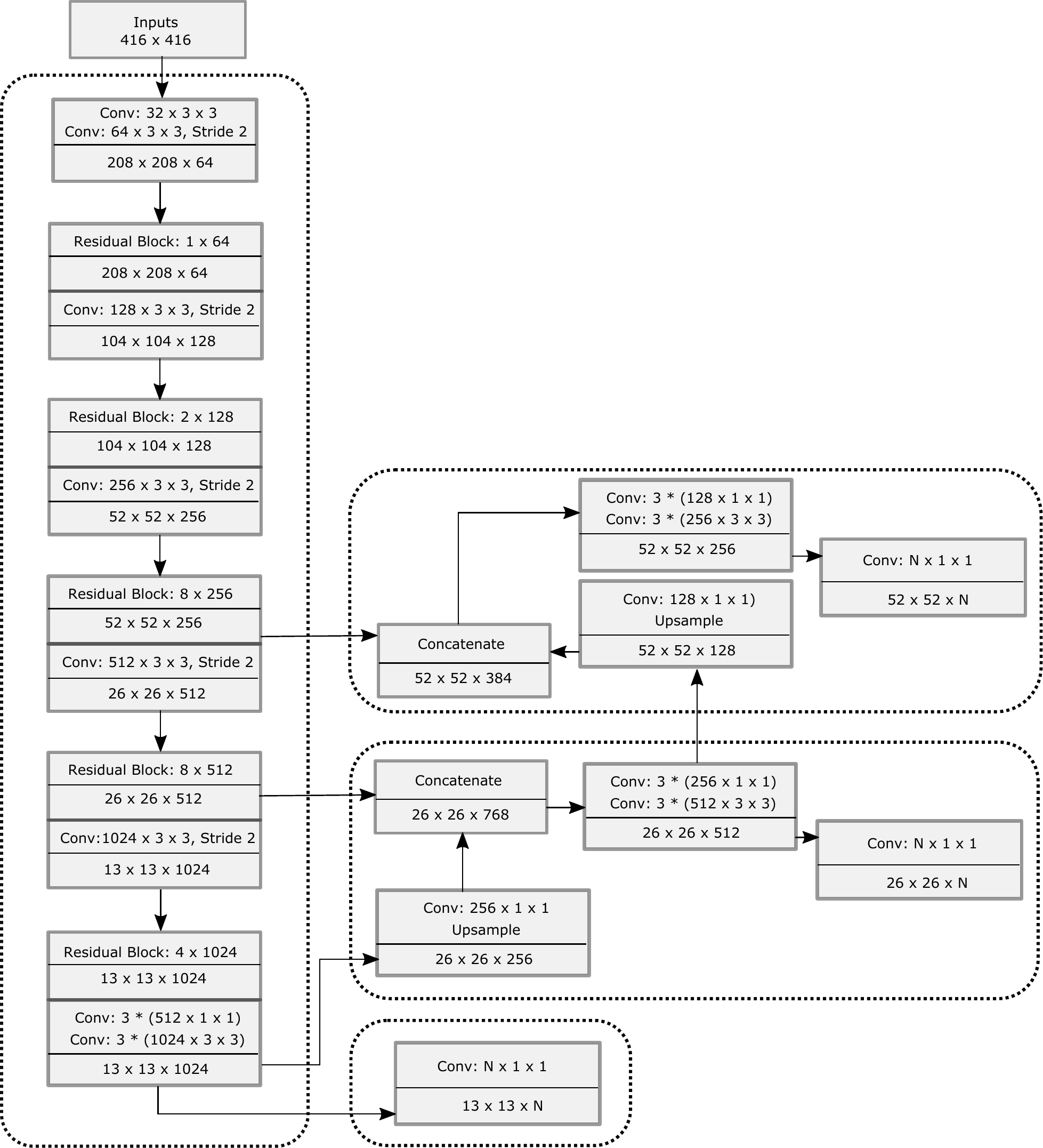}

    \caption{ Network architecture of Keypoint proposal network (YOLOv3 \cite{yolov3}). $N$ denotes the number of outputs from each grid cell.}
\label{fig:backbone_arch}
\end{figure*}

\begin{figure*}[t]
\centering
          \includegraphics[width=\textwidth]{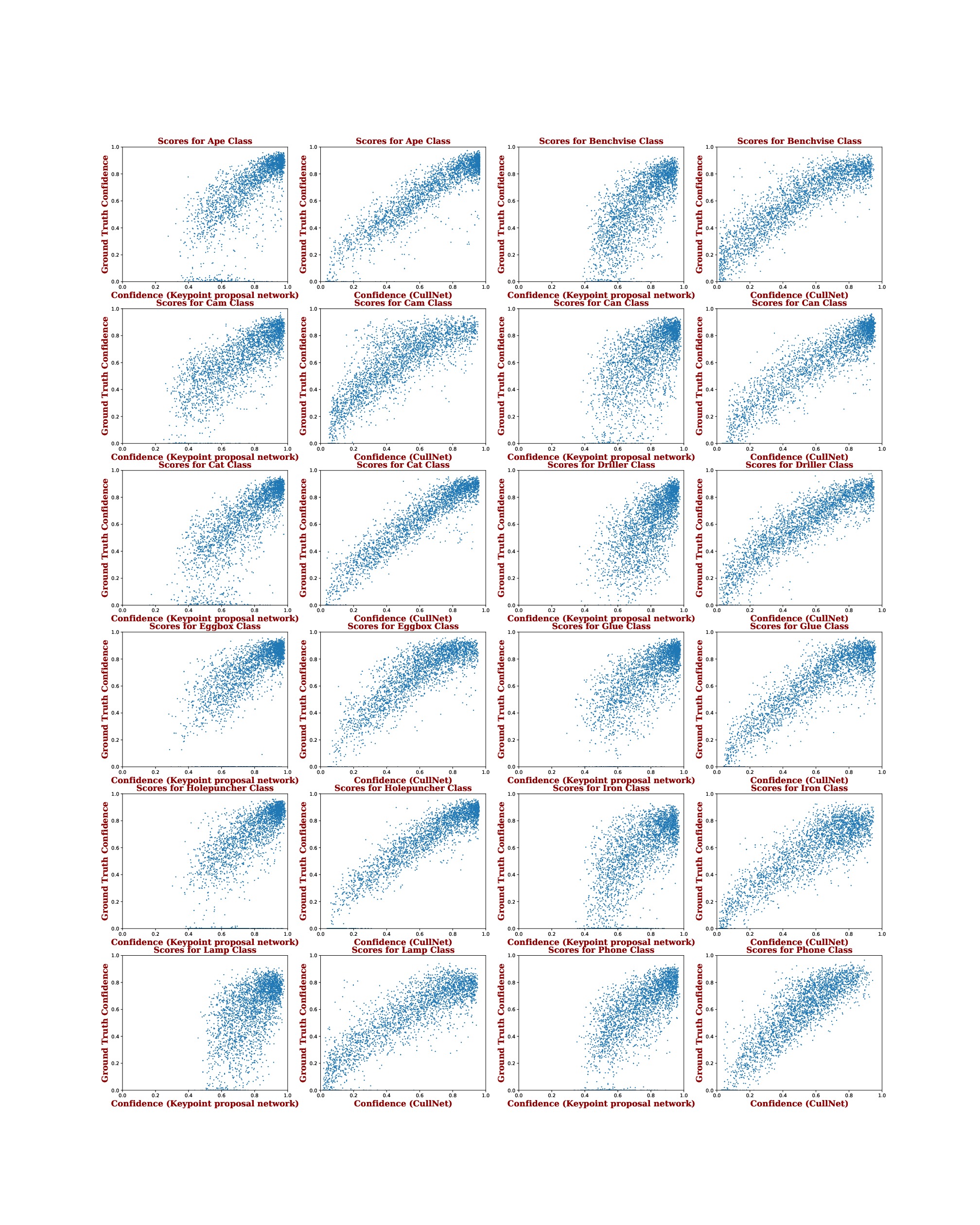}
    \vspace*{-2cm}
          
    \caption{ Comparisons of pose proposal confidence output of Keypoint proposal network and CullNet for different classes of LINEMOD dataset.}
\label{fig:ConfidencePlot_supp}
\end{figure*}

\paragraph{Confidence Plots}
We show comparisons of pose proposal confidence output of Keypoint proposal network and CullNet for all classes\footnote{For 'Duck' class, we show these plots in the main paper.} of LINEMOD dataset in Figure \ref{fig:ConfidencePlot_supp}.


\fi
\twocolumn
{\small
\bibliographystyle{ieee}
\bibliography{references}
}

\end{document}